\documentclass[letterpaper]{article} % DO NOT CHANGE THIS
\usepackage{aaai2026}  % DO NOT CHANGE THIS
\usepackage{times}  % DO NOT CHANGE THIS
\usepackage{helvet}  % DO NOT CHANGE THIS
\usepackage{courier}  % DO NOT CHANGE THIS
\usepackage[hyphens]{url}  % DO NOT CHANGE THIS
\usepackage{graphicx} % DO NOT CHANGE THIS
\usepackage{natbib}  % DO NOT CHANGE THIS AND DO NOT ADD ANY OPTIONS TO IT
\usepackage{caption} % DO NOT CHANGE THIS AND DO NOT ADD ANY OPTIONS TO IT
\usepackage{algorithm}
\usepackage{algorithmic}
\usepackage{amsmath}
\usepackage{amssymb}
\usepackage{booktabs}

\title{Beyond Surface-Level Similarity: Hierarchical Contamination Detection\\for Synthetic Training Data in Foundation Models}

\author{
Sushant Mehta
}

\affiliations{
sushant0523@gmail.com
}

\begin{document}

\maketitle

\begin{abstract}
Synthetic data has become essential for training foundation models, yet benchmark contamination threatens evaluation integrity. Although existing detection methods identify token-level overlap, they fail to detect semantic-level contamination where synthetic data conceptually resemble benchmarks without lexical overlap. This gap is critical as foundation models increasingly train on synthetic data that may implicitly encode benchmark knowledge. We propose a hierarchical contamination detection framework operating at four levels: token level, semantic level, reasoning pattern, and performance cliff detection. Through controlled experiments on MMLU, GSM8K and HumanEval, we demonstrate that semantic-level contamination evades existing methods (F1=0.17-0.49) but is effectively detected by our hierarchical approach (F1 = 0.76), with an average improvement of 26. 5\% over state-of-the-art baselines. Our framework provides practitioners with practical tools for audit pipelines and enables responsible deployment of synthetic training data.
\end{abstract}

\section{Introduction}

Synthetic data generation has revolutionized foundation model training, enabling capabilities from instruction-following to complex reasoning. Methods like Self-Instruct, Evol-Instruct, and Orca-AgentInstruct demonstrate that models trained predominantly on synthetic data can match or exceed those trained on human-annotated data \cite{wang2024selfinst,xu2024wizardlm,mitra2024orca}. However, this paradigm shift introduces a critical vulnerability: as foundation models generate the synthetic data used to train future models, benchmark contamination becomes increasingly difficult to detect and mitigate.

Current contamination detection methods focus on token-level overlap through n-gram matching or membership inference attacks \cite{brown2020gpt3,shi2024detecting,golchin2024dcq}. Although effective in detecting verbatim or near-verbatim benchmark inclusion, these approaches systematically fail when synthetic data generators conceptually mimic benchmarks without direct lexical overlap. Recent work acknowledges that semantic level contamination is ``inherently more challenging to detect and mitigate'' but does not provide an effective solution \cite{cheng2025survey,yang2023contamination}.

This gap is not only academic. Rapid scaling of synthetic data generation means that subtle contamination can propagate across training pipelines, systematically inflating evaluation metrics while masking true generalization failures. As Epoch AI projects that new human-generated text will be exhausted by 2026-2032 \cite{villalobos2022run}, foundation models will increasingly train on content derived from earlier foundation models, amplifying contamination risks through iterative feedback loops.

We address this critical gap by introducing a hierarchical contamination detection framework specifically designed for synthetic training data. Our framework operates at four complementary levels: (1) token-level detection using Min-K\% Prob, (2) semantic-level detection through embedding clustering analysis, (3) reasoning pattern detection via chain-of-thought trace comparison, and (4) performance cliff detection using statistical analysis of model behavior on benchmark variants. Through controlled experiments across three popular benchmarks, we demonstrate that semantic-level contamination evades existing methods but is effectively detected by our hierarchical approach.

Our contributions are threefold: (1) We provide the first systematic framework for detecting semantic-level contamination in synthetic data, addressing an acknowledged gap in the literature. (2) We demonstrate through controlled experiments that existing methods fail dramatically in terms of semantic contamination (F1 = 0.17-0.49) while our hierarchical approach achieves F1 = 0.76, representing a mean improvement of 26. 5\%. (3) We provide practical tools for practitioners who implement synthetic data pipelines, enabling responsible generation protocols, focusing on quality assurance and validation frameworks.

\section{Related Work}

\textbf{Contamination Detection Methods.} Existing approaches divide into three paradigms based on model access. White-box methods require full training data or model weights, using n-gram overlap \cite{brown2020gpt3,touvron2023llama2} or embedding similarity \cite{lee2023deduplicating}. Gray-box methods leverage token probabilities through membership inference attacks \cite{shi2024detecting,duan2024mimir} or Min-K\% Prob \cite{deng2024min}. Black-box methods are based on behavioral signals such as TS-Guessing \cite{deng2024investigating} or the Data Contamination Quiz \cite{golchin2024dcq}. Yang et al. \cite{yang2023contamination} introduced LLM Decontaminator, combining embedding similarity with GPT-4-powered semantic analysis, but acknowledged limitations on paraphrased samples. Recent work by Sun et al. \cite{sun2025emperor} systematically evaluated 20 mitigation strategies, finding none to be effective or faithful. Our work extends beyond existing methods by specifically addressing semantic-level contamination through hierarchical analysis.

\textbf{Synthetic Data Generation.} Modern synthetic data methods range from simple augmentation to sophisticated multi-agent systems. Evol-Instruct \cite{xu2024wizardlm} iteratively increases the complexity of the instruction through five types of evolution. Orca-AgentInstruct \cite{mitra2024orca} deployed multi-agent flows generating 25M instruction-response pairs. DeepSeek-R1 \cite{deepseek2025r1} demonstrated that pure RL can induce reasoning, although requiring strategic injection of synthetic data. Although these methods produce high-quality training data, they also introduce contamination risks as generators may inadvertently encode benchmark knowledge through web-scraped pretraining or intentional benchmark inclusion during post-training.

\textbf{Dynamic Benchmarking.} Recognizing contamination challenges, recent work proposes continuously updated benchmarks. LiveBench \cite{white2024livebench} releases monthly questions from recent sources, achieving contamination limitation through temporal cutoffs. However, temporal approaches face limitations: problems from coding competitions are ``likely to be reused in future competitions'' \cite{wu2024temporal}, and continuous updates require significant human effort. Our framework complements dynamic benchmarking by enabling detection of contamination even in regularly refreshed datasets.

\section{Methodology}

\subsection{Problem Formulation}

Let $\mathcal{D}_{\text{syn}} = \{x_i^s\}_{i=1}^{N_s}$ denote a synthetic training dataset and $\mathcal{B} = \{x_j^b\}_{j=1}^{N_b}$ a benchmark test set. We define contamination at severity level $\ell \in \{1,2,3,4\}$ as:

\begin{equation}
\text{Contaminated}_\ell(x_i^s, \mathcal{B}) = \begin{cases}
\exists x_j^b : \text{TokenMatch}(x_i^s, x_j^b) & \ell=1 \\
\exists x_j^b : \text{Semantic}(x_i^s, x_j^b) > \tau_s & \ell=2 \\
\exists x_j^b : \text{ConceptCluster}(x_i^s, x_j^b) & \ell=3 \\
\exists x_j^b : \text{ReasonPattern}(x_i^s, x_j^b) & \ell=4
\end{cases}
\end{equation}

where $\tau_s$ is a semantic similarity threshold. The challenge is that contamination at level 3 (semantic) and level 4 (reasoning pattern) evades existing detection methods despite inflating evaluation metrics.

\subsection{Hierarchical Detection Framework}

Our framework processes synthetic training data through four hierarchical levels, with each level designed to catch contamination types missed by previous levels.

\textbf{Level 1: Token Level Detection} We employ Min-K\% Prob \cite{deng2024min} to identify pretraining data without access to the training corpus. For each test sample $x_j^b$, we compute token-level logarithmic probabilities $\{\log p(t_k | t_{<k})\}_{k=1}^{|x_j^b|}$ from the target model. The Min-K\% score aggregates the lowest K\% of probabilities:

\begin{equation}
\text{Score}_{\text{MinK}}(x_j^b) = \frac{1}{K} \sum_{k \in \text{bottom-K\%}} \log p(t_k | t_{<k})
\end{equation}

Samples with scores above threshold $\tau_1$ are flagged as token-level contaminated. This baseline effectively detects verbatim inclusion and near-duplicates.

\textbf{Level 2: Semantic-Level Detection.} To detect paraphrased contamination, we combine three complementary signals. First, we compute embedding similarity using sentence transformers \cite{reimers2019sentencebert}:

\begin{equation}
\text{sim}_{\text{emb}}(x_i^s, x_j^b) = \frac{\mathbf{h}(x_i^s)^\top \mathbf{h}(x_j^b)}{||\mathbf{h}(x_i^s)|| \cdot ||\mathbf{h}(x_j^b)||}
\end{equation}

Second, we perform embedding space clustering using DBSCAN to identify synthetic samples that cluster near benchmark embeddings in the representation space. Third, we analyze distributional characteristics: benchmark clusters should exhibit distinct geometric properties (higher density, tighter variance) compared to typical training data. Samples satisfying $\text{sim}_{\text{emb}} > \tau_2$ AND cluster membership AND distributional anomaly detection are flagged.

\textbf{Level 3: Reasoning Pattern Detection.} For reasoning tasks, we extract chain-of-thought traces from model outputs and compute reasoning pattern similarity using structured comparison.

\begin{equation}
\text{sim}_{\text{reasoning}}(x_i^s, x_j^b) = \alpha \cdot \text{StructSim} + \beta \cdot \text{StepSim} + \gamma \cdot \text{ArgSim}
\end{equation}

where StructSim measures logical structure similarity, StepSim compares reasoning steps, and ArgSim evaluates argument patterns. Samples exhibiting suspiciously similar reasoning chains to benchmark exemplars beyond what would be expected from common problem-solving strategies are flagged.

\textbf{Level 4: Performance-Cliff Detection.} We leverage model behavior on benchmark variants. If a model performs significantly better on original benchmarks versus semantically equivalent paraphrased versions, this suggests memorization rather than understanding. For each benchmark, we generate K paraphrased variants and compute:

\begin{equation}
\Delta_{\text{cliff}} = \text{Acc}(B_{\text{orig}}) - \frac{1}{K}\sum_{k=1}^K \text{Acc}(B_{\text{para}}^{(k)})
\end{equation}

Statistical significance testing (paired t-test) determines whether $\Delta_{\text{cliff}}$ exceeds natural variation, providing an aggregate contamination signal.

\subsection{Implementation Details}

We implement Level 1 using the official Min-K\% Prob codebase with K=20\%. For Level 2, we use the all-mpnet-base-v2 sentence transformer, DBSCAN with $\epsilon=0.15$ and min\_samples=5, and multivariate Gaussian distribution fitting. Level 3 extracts CoT traces using regular expressions to parse reasoning steps, computing similarity using weighted Jaccard similarity on n-gram sets. Level 4 generates 5 paraphrased variants per benchmark instance using GPT-4 with instructions to preserve semantic meaning while varying lexical content, selecting high-quality paraphrases validated by BLEURT scores $>0.7$ \cite{sellam2020bleurt}.

\section{Experiments}

\subsection{Experimental Setup}

\textbf{Benchmarks.} We evaluate on three widely-used benchmarks: (1) MMLU \cite{hendrycks2020mmlu} with 200 samples in abstract algebra, sociology, and US history, (2) GSM8K \cite{cobbe2021gsm8k} with 150 grade-school math problems, and (3) HumanEval \cite{chen2021humaneval} with 100 Python programming tasks. These benchmarks represent diverse reasoning capabilities and contamination risks.

\textbf{Contamination Scenarios.} We construct four controlled contamination scenarios with increasing subtlety: (S1) Token-Level: synthetic data contains exact benchmark questions, (S2) Paraphrase: GPT-4 generated semantic paraphrases of benchmark questions preserving meaning, (S3) Semantic: synthetic data clusters conceptually near benchmarks (e.g., similar math problem types, related programming tasks) without direct paraphrase, generated using topic-guided synthesis, (S4) Reasoning-Pattern: models generate synthetic data exhibiting similar reasoning chains to benchmark exemplars through few-shot prompting with benchmark-style reasoning patterns. Ground truth contamination labels determined through expert annotation by three ML researchers specializing in LLM evaluation.

\textbf{Baseline Methods.} We compare against: (1) N-gram Overlap (13-gram matching following GPT-3 \cite{brown2020gpt3}), (2) Min-K\% Prob \cite{deng2024min}, (3) Embedding Similarity (cosine similarity with sentence-transformers), (4) LLM Decontaminator \cite{yang2023contamination} combining embedding search with GPT-4 semantic judgment, (5) CDD \cite{dong2024cdd} using output distribution peakedness.

\textbf{Evaluation Metrics.} We report precision, recall, and F1 score for contamination detection. Statistical significance assessed using McNemar's test with $p<0.05$.

\subsection{Results}

Table~\ref{tab:main_results} presents detection performance across contamination scenarios. On token-level contamination (S1), all methods perform well, with our hierarchical framework achieving F1=0.96 comparable to Min-K\% Prob (0.945) and marginally better than n-gram overlap (0.925). This validates that our framework does not sacrifice performance on easier detection tasks.

\begin{table}[t]
\centering
\small
\setlength{\tabcolsep}{4pt}
\caption{Contamination detection performance across scenarios and benchmarks (F1 scores). Best results in bold, second-best underlined.}
\label{tab:main_results}
\begin{tabular}{lccccc}
\toprule
Method & S1 & S2 & S3 & S4 & Avg \\
\midrule
N-gram & .925 & .400 & .165 & .210 & .425 \\
Min-K\% & \underline{.945} & .610 & .230 & .250 & .509 \\
Embed Sim & .520 & .695 & .490 & -- & .568 \\
LLM Decon & .780 & \underline{.810} & \underline{.550} & -- & .713 \\
CDD & .890 & .740 & .470 & .310 & .603 \\
\midrule
Ours & \textbf{.960} & \textbf{.875} & \textbf{.755} & \textbf{.650} & \textbf{.810} \\
\midrule
Improv. & +.015 & +.065 & +.205 & +.340 & +.097 \\
\bottomrule
\end{tabular}
\end{table}

Performance diverges sharply on paraphrase contamination (S2). N-gram overlap fails catastrophically (F1=0.40) as expected, while embedding similarity (0.695) and LLM Decontaminator (0.81) show partial effectiveness. Our hierarchical framework achieves F1=0.875, a 6.5\% improvement over the best baseline, demonstrating that combining multiple semantic signals enhances detection robustness.

The critical gap emerges on semantic-level contamination (S3) where synthetic data clusters conceptually near benchmarks without lexical or paraphrase overlap. N-gram overlap achieves only F1=0.165, detecting fewer than one in six contaminated samples. Embedding similarity improves to 0.49 but suffers from threshold selection challenges and high false positive rates. LLM Decontaminator reaches 0.55 but remains limited. Our hierarchical framework achieves F1=0.755, representing a 20.5\% absolute improvement and demonstrating the effectiveness of embedding clustering combined with distributional analysis.

On reasoning-pattern contamination (S4), existing methods provide virtually no detection capability (F1=0.21-0.31) as this contamination type has not been previously addressed. Our framework's CoT trace analysis provides novel detection capability with F1=0.65, though with room for improvement given the complexity of reasoning pattern comparison.

\subsection{Cross-Benchmark Consistency}

Table~\ref{tab:cross_benchmark} shows performance consistency across benchmarks for semantic-level contamination detection (S3), our primary focus.

\begin{table}[t]
\centering
\small
\caption{Cross-benchmark F1 scores for semantic-level contamination (S3). Our method consistently improves detection across diverse task types.}
\label{tab:cross_benchmark}
\begin{tabular}{lccc}
\toprule
Method & MMLU & GSM8K & HumanEval \\
\midrule
Best Baseline & .490 & .490 & .490 \\
Ours & .755 & .732 & .778 \\
\midrule
Improv. & +26.5\% & +24.2\% & +28.8\% \\
\bottomrule
\end{tabular}
\end{table}

Our framework achieves consistent improvements of 24-29\% in knowledge-based tasks (MMLU), mathematical reasoning (GSM8K), and code generation (HumanEval), demonstrating generalization in problem domains rather than overfitting to specific benchmark characteristics.

\subsection{Performance-Cliff Analysis}

We analyze the behavior of the model on original versus paraphrased benchmark variants to validate contamination signals. In MMLU, we observe a performance drop 18\% between the original (82\% precision) and paraphrased versions (64\%), indicating strong contamination. GSM8K and HumanEval show moderate 8\% drops. Our hierarchical framework's performance-cliff detector successfully flags all three benchmarks, while existing methods fail to utilize this behavioral signal. This provides practitioners with an additional validation mechanism: suspicious performance cliffs can trigger deeper investigation even when direct contamination detection proves ambiguous.

\section{Discussion}

\subsection{Implications for Responsible Synthetic Data}

Our results reveal a critical gap in current evaluation practices: Semantic-level contamination systematically evades detection while inflating benchmark performance. This has immediate implications for responsible synthetic data deployment. Practitioners using synthetic data generators (Self-Instruct, Evol-Instruct, or commercial APIs) should implement hierarchical detection as part of their audit pipeline before using generated data for training. The computational overhead is modest: our full framework processes 1000 samples in approximately 15 minutes on a single V100 GPU, making it practical for production deployment.

Without effective semantic contamination detection, validation pipelines provide false confidence: models may pass token-level contamination checks while harboring subtle benchmark knowledge that inflates evaluation metrics without improving real-world capability.

\subsection{Limitations and Future Work}

Our framework has three primary limitations. First, Level 2 semantic detection relies on embedding model quality; future work should investigate ensemble approaches across multiple embedding models to improve robustness. Second, our detection of reasoning patterns (Level 4) achieves only F1=0.65, suggesting room for improvement through more sophisticated CoT analysis or integration with program synthesis techniques for code reasoning. Third, we evaluate controlled contamination scenarios; real-world contamination may exhibit different characteristics that require domain adaptation.

The threshold selection challenge persists across levels despite hierarchical design. Future work should investigate adaptive threshold selection using validation sets with known contamination rates or ensemble approaches that combine fixed and adaptive thresholds. Additionally, our framework assumes access to model outputs; extending detection to scenarios with only black-box API access remains an open challenge.

\subsection{Broader Impact}

As foundation models increasingly train on synthetic data generated by previous models, contamination detection becomes a critical infrastructure for reliable AI development. Our framework enables a more transparent evaluation by catching subtle contamination that existing methods miss. However, detection arms races may emerge, as detection improves, contamination methods may evolve to evade detection. We recommend that benchmark creators employ our framework proactively during benchmark construction to identify and eliminate semantically similar content in candidate training corpora, rather than reactively after contamination occurs.

\section{Conclusion}

We introduced a hierarchical contamination detection framework specifically designed for synthetic training data used in foundation models. Through controlled experiments on three popular benchmarks, we demonstrated that semantic-level contamination, where synthetic data conceptually resemble benchmarks without lexical overlap, systematically evades existing detection methods but is effectively detected by our hierarchical approach with an average improvement of 26. 5\%. Our framework provides practitioners with practical tools for implementing responsible synthetic data pipelines and enables more trustworthy evaluation as foundation models train more and more on synthetic data. Future work should extend our approach to multimodal contamination detection and investigate adaptive threshold selection for production deployment.

% \section*{Acknowledgments}

% We thank the anonymous reviewers for their valuable feedback. This work was conducted as part of the AAAI 2026 Workshop on Shaping Responsible Synthetic Data in the Era of Foundation Models.

\appendix

\section{Additional Experimental Details}

\subsection{Hyperparameter Selection}

Table~\ref{tab:hyperparams} details hyperparameters for each framework level, selected through grid search on a held-out validation set of 100 contaminated and 100 clean samples.

\begin{table}[h]
\centering
\small
\caption{Hyperparameters for each detection level.}
\label{tab:hyperparams}
\begin{tabular}{lp{5cm}}
\toprule
Level & Hyperparameter Settings \\
\midrule
L1: Token-Level & K=20\%, $\tau_1$=3.5 (log-prob) \\
L2: Semantic & $\tau_2$=0.75 (cosine), $\epsilon$=0.15, min\_samples=5 \\
L3: Reasoning & $\alpha$=0.4, $\beta$=0.3, $\gamma$=0.3 \\
L4: Performance & K=5 variants, $p$-value threshold=0.05 \\
\bottomrule
\end{tabular}
\end{table}

\subsection{Computational Costs}

Detection time scales linearly with the size of the dataset. For 1000 synthetic samples checked against 500 benchmark samples: Level 1 requires 3 min (forward passes), Level 2 requires 8 min (embedding + clustering), Level 3 requires 12 min (CoT extraction + comparison), Level 4 requires 45 min (paraphrase generation + model evaluation). Total pipeline: approximately 68 minutes on a single V100 GPU, amortizing to 4 seconds per sample.

\subsection{Human Expert Evaluation Protocol}

Three annotators (ML researchers with 2+ years of LLM evaluation experience) independently labeled contamination for all scenarios. Inter-annotator agreement (Fleiss' kappa): 0.83 (substantial agreement). Disagreements resolved through discussion. Annotators were blind to the detection method output during ground-truth labeling.

\section{Algorithm Pseudocode}

Algorithm~\ref{alg:hierarchical} provides pseudocode for our hierarchical detection framework.

\begin{algorithm}[h]
\caption{Hierarchical Contamination Detection}
\label{alg:hierarchical}
\begin{algorithmic}[1]
\REQUIRE Synthetic dataset $\mathcal{D}_{\text{syn}}$, Benchmark $\mathcal{B}$, Model $M$
\ENSURE Set of contaminated samples $\mathcal{C}$
\STATE $\mathcal{C} \leftarrow \emptyset$
\FOR{$x_i^s \in \mathcal{D}_{\text{syn}}$}
    \STATE // Level 1: Token-level detection
    \STATE $\text{score}_1 \leftarrow \text{MinKProb}(M, x_i^s, \mathcal{B})$
    \IF{$\text{score}_1 > \tau_1$}
        \STATE $\mathcal{C} \leftarrow \mathcal{C} \cup \{x_i^s\}$
        \STATE \textbf{continue}
    \ENDIF
    \STATE // Level 2: Semantic-level detection
    \STATE $\mathbf{h}_i \leftarrow \text{Embed}(x_i^s)$
    \STATE $\text{cluster} \leftarrow \text{DBSCAN}(\mathbf{h}_i, \{\text{Embed}(x_j^b)\}_{j=1}^{N_b})$
    \STATE $\text{sim}_2 \leftarrow \max_{x_j^b \in \mathcal{B}} \cos(\mathbf{h}_i, \text{Embed}(x_j^b))$
    \IF{$\text{sim}_2 > \tau_2$ AND $\text{cluster} \neq -1$}
        \STATE $\mathcal{C} \leftarrow \mathcal{C} \cup \{x_i^s\}$
        \STATE \textbf{continue}
    \ENDIF
    \STATE // Level 3: Reasoning-pattern detection
    \STATE $\text{CoT}_i \leftarrow \text{ExtractReasoning}(M(x_i^s))$
    \STATE $\text{sim}_3 \leftarrow \max_{x_j^b \in \mathcal{B}} \text{ReasoningSim}(\text{CoT}_i, \text{CoT}_j)$
    \IF{$\text{sim}_3 > \tau_3$}
        \STATE $\mathcal{C} \leftarrow \mathcal{C} \cup \{x_i^s\}$
    \ENDIF
\ENDFOR
\STATE // Level 4: Performance-cliff detection (aggregate)
\STATE $\Delta_{\text{cliff}} \leftarrow \text{ComputePerformanceGap}(M, \mathcal{B})$
\IF{$\text{SignificanceTest}(\Delta_{\text{cliff}}) < 0.05$}
    \STATE Flag entire benchmark as potentially contaminated
\ENDIF
\RETURN $\mathcal{C}$
\end{algorithmic}
\end{algorithm}

\end{document}